# AI-Driven Personalized Learning: Predicting Academic Performance Through Leadership Personality Traits

Nitsa J Herzog[1,*], Rejwan Bin Sulaiman[2], David J Herzog[3], Rose Fong[4]

[1] Northumbria University London Campus, Department of Environmental Engineering, London, UK
[3] Ulster University, London Campus, Department of Computing, London, UK
* Correspondence: nitsa.herzog@northumbria.ac.uk.

**Abstract**

The study explores the potential of AI technologies in personalized learning, suggesting the prediction of academic success through leadership personality traits and machine learning modelling. The primary data were obtained from 129 master's students in the Environmental Engineering Department, who underwent five leadership personality tests with 23 characteristics. Students used self-assessment tools that included Personality Insight, Workplace Culture, Motivation at Work, Management Skills, and Emotion Control tests. The test results were combined with the average grade obtained from academic reports. The study employed exploratory data analysis and correlation analysis. Feature selection utilized Pearson correlation coefficients of personality traits. The average grades were separated into three categories: "Fail", "Pass", and "Excellent". The modelling process was performed by tuning seven ML algorithms, such as Support Vector Machine, Logistic Regression, K-Nearest Neighbors, Decision Tree, Gradient Boosting, Random Forest, XGBoost and LightGBM. The highest predictive performance was achieved with the Random Forest classifier, which yielded an accuracy of 87.50% for the model incorporating 17 personality trait features and the leadership mark feature, and an accuracy of 85.71% for the model excluding this feature. In this way, the study offers an additional opportunity to identify students' strengths and weaknesses at an early stage of their education process and select the most suitable strategies for personalized learning.

## 1. Introduction

The popularity of customized education services and personalized learning/teaching approaches has attracted considerable attention in recent years. AI-driven learning is going to replace traditional methods [1]. The prediction of outcomes from the usage of AI technologies and their effects on the educational sector are actively evaluated [2]. Personalized learning heavily depends on the individual's character traits that reflect specific behavioral patterns [3]. Understanding the individual traits that impact academic performance can significantly enhance students' learning experiences, increase their motivation, and improve their engagement in learning activities.

The modern world relies heavily on digitalization and teamwork. Digital leaders can be trained according to organizational needs. Many educational bodies include leadership training in their curriculum. Personal leadership traits can be identified through specialized career tests and strengthened using personal development strategies and approaches. Previous research evaluating the effect of personality types on academic achievement found strong correlations between them. However, AI technologies have opened up additional opportunities to utilize this knowledge in personalized teaching and learning processes [4]. The motivation for this study was to investigate the dependencies between leadership traits, which were not explored before, and student performance by evaluating a comprehensive set of self-assessment results and creating an AI-driven predictive model.

The study aimed to answer several research questions:
- How do leadership personality traits correlate with students' academic performance?
- Which leadership personality traits appear most influential in predicting higher student grades?
- Can a machine learning model accurately predict students' grade categories based on leadership personality traits?
- How can universities use insights from this study to develop personalized academic strategies tailored to students' strengths and personality profiles?

The research was focused on completing the following objectives:
- Identify a diverse group of students from different programs in the Environmental Engineering (EE) department.
- Explore results from self-assessment, including Personality Insight test, Workplace Culture, Motivation at Work, Emotional Control, and Management Skills.
- Align test results with students' average academic mark across several modules.
- Analyze the dataset using quantitative statistical methods.
- Determine which personality traits appear to be the most influential for academic success.
- Create a machine learning model that can predict grade categories, such as "Pass", "Fail", and "Distinction".
- Formulate recommendations for the university to enhance student outcomes and help to develop personalized academic strategies based on individual strengths and personality profiles.

The article consists of 5 sections. Section 2, "Related Works", represents the determinant of academic performance in the light of modern leadership theories, discusses the previous studies that explored personality types and students' achievements, and highlights AI-based studies targeting the predictions of students' grades. Section 3, "Materials and Methods", describes the data collection procedures and discusses the study's design. The following "Results" section focuses on data analytical parts, feature selection and modelling stages. The "Discussion and Conclusion" section discusses the teaching strategies towards personalized teaching and learning using ML predictions.

## 2. Related Works

This literature review explores modern leadership trait theories, determinants of student academic success, the role of personality in academic achievement, self-assessment as a measurement tool, and the latest advancements in AI for grade prediction and personality-based performance prediction. At the end of the literature review, a research gap is identified in integrating AI-based personality assessments with leadership theories to predict student performance in educational contexts.

*2.1. The Modern Leadership Trait Theories Overview*

In the early 20th century, as leadership theories began to emerge, researchers believed that leadership was an innate quality—that leaders were born, not made. This belief was formalized in the Great Man Theory by Carlyle in his book "On Heroes, Hero-Worship and the Heroic in History" published in 1894 [5]. The theory was supported by examples of leaders from privileged backgrounds, including those with elite education and wealthy families. However, following the Industrial and later the Technological Revolutions, the theory's relevance declined. It was later re-evaluated by Cawthon in "Leadership: The Great Man Theory Revisited" in 1996 [6].

Because the Great Man Theory lacked empirical support, a new approach emerged— the Trait Theory of Leadership. While it shared the idea that certain individuals possess qualities that make them more effective leaders, it differed in two keyways: it identified specific traits and skills that contribute to leadership success, and it proposed that these attributes can be developed through education and experience. Subsequently, researchers shifted their focus to behavioral and situational theories, which emphasized the influence of personal traits in relation to specific contexts [7].

In fact, Kirkpatrick and Locke in 1991 identified six traits [8] that distinguish effective leaders from others:
- Drive (ambition and energy),
- Desire to lead (motivational inclination toward leadership roles),
- Honesty and integrity (trustworthiness and ethics),
- Self-confidence (confidence in one's abilities),
- Cognitive ability (intelligence and analytical skills),
- Knowledge of the field (expertise in the relevant domain).

Modern leadership trait theory has evolved significantly from early "great man" theories to more nuanced, empirically supported frameworks. Contemporary research acknowledges that effective leadership stems from a complex interplay of traits, behaviors, and situational factors [9]. The Big Five personality model has become particularly influential in leadership research, with meta-analyses demonstrating consistent relationships between personality dimensions and leadership effectiveness [10]. Current literature identifies several key trait categories that consistently correlate with leadership effectiveness:
- Cognitive Abilities: Intelligence, particularly emotional intelligence, has emerged as a critical leadership trait. Goleman's [11] emotional intelligence framework encompasses self-awareness, self-regulation, motivation, empathy, and social skills. Meta-analytic evidence supports moderate to strong correlations between cognitive abilities and leadership performance [12].
- Personality Dimensions: The Five-Factor Model provides a robust framework for understanding leadership-relevant personality traits. Extraversion shows the strongest correlation with leadership emergence and effectiveness, followed by conscientiousness and openness to experience [12]. Neuroticism typically shows negative correlations with leadership effectiveness.
- Motivational Factors: Achievement motivation, power motivation, and leadership motivation have been identified as crucial drivers of leadership behavior. McClelland's research [13] on implicit motives demonstrates that effective leaders often exhibit high power motivation combined with impulse control.

Recent developments in leadership trait theory include the integration of character strengths [14], authentic leadership traits [15], and servant leadership characteristics [16]. These approaches emphasize moral dimensions and follower development as core leadership competencies.

*2.2. Review of Established Determinants/Academic Performance Factors of Student Success*

A complex interplay of cognitive, non-cognitive, and environmental factors influences academic performance. Cognitive factors, particularly academic ability and intelligence are widely recognized as primary predictors of success in educational settings. Richardson et al.'s [17] meta-analysis demonstrated that prior academic performance is the strongest correlation of future GPA, with a correlation coefficient of 0.34. General intelligence and specific academic competencies consistently predict outcomes across all levels of education. However, non-cognitive traits also play a crucial role. Among personality dimensions, conscientiousness has emerged as a reliable predictor of academic success, with Poropat's [18] meta-analysis showing its effect sizes rival those of intelligence. Additionally, self-regulation and effective study strategies, as outlined in Zimmerman's theory [19], enhance academic outcomes by promoting goal setting and strategic learning behaviors. Motivation, particularly mastery goal orientation as described by Dweck [20], is another key determinant, leading to sustained academic achievement. Environmental and social variables further influence educational attainment. Socioeconomic status impacts academic success through access to resources and cultural capital [21] while social support systems, including family, peers, and faculty, have been shown to improve retention and performance, as emphasized in Tinto's model [22]. Institutional characteristics such as teaching quality and support services also contribute meaningfully to student outcomes [23]. Together, these factors underscore the multifaceted nature of academic achievement and emphasize the importance of considering both individual traits and contextual influences.

*2.3. Previous Studies on Personality and Academic Achievement*

Personality traits, especially conscientiousness, are strong predictors of academic performance, with consistent positive correlations due to factors like time management and persistence [24]. Openness to experience also supports academic success in intellectually demanding contexts [25], while neuroticism tends to hinder performance due to anxiety [26]. Beyond the Big Five, academic self-efficacy significantly predicts performance and persistence [27], whereas the role of grit is debated, with limited value beyond conscientiousness [28]. Cultural and contextual differences influence how personality traits relate to academic outcomes, varying across cultures, disciplines, and assessment types [25].

*2.4. Self-Assessment as a Measurement Tool and Its Reliability and Validity in Educational Contexts*

Self-assessment, grounded in metacognitive theory, plays a central role in self-regulated learning by enabling students to evaluate their learning and performance [19]. Its effectiveness relies heavily on metacognitive awareness and the ability to accurately calibrate self-perceptions. In terms of reliability, self-report personality measures typically demonstrate good internal consistency, with Cronbach's alpha values ranging from 0.70 to 0.90 in tools like the NEO-PI-R and the Big Five Inventory [29]. These assessments also show moderate to high test-retest reliability, particularly for core personality traits, with stability increasing with age and correlations above 0.70 over several years [30]. Furthermore, the inter-rater agreement between self-reports and observer ratings typically falls between 0.40 and 0.60, indicating moderate convergent validity and the distinct perspective that self-assessments offer [31].

However, self-assessment is limited by several validity concerns. One major issue is social desirability bias, where individuals present themselves in a more favorable light, thereby reducing the accuracy of their responses, particularly for traits like conscientiousness [32]. Additionally, people often engage in self-enhancement, overestimating their abilities and traits, although some self-enhancement can positively relate to psychological well-being and performance [33]. Accuracy also varies by domain; students tend to be more precise in assessing their performance on concrete tasks than on complex or ambiguous ones [34].

To improve the validity of self-assessment, training and scaffolding strategies, such as structured rubrics, exemplars, and guided practice, have proven effective in enhancing accuracy [35]. Technology can further support this process by offering immediate feedback and calibrated comparisons to objective measures. Digital platforms and adaptive systems help refine self-assessment accuracy by aligning subjective evaluations with actual performance outcomes [36].

*2.5. AI in Education*

Advancements in artificial intelligence and machine learning have significantly enhanced educational data mining and the prediction of student performance. Machine learning algorithms can process large-scale datasets comprising student demographics, behavioral patterns, and learning analytics to identify trends and predict academic outcomes with high accuracy. Supervised learning models such as random forests, support vector machines, and gradient boosting techniques outperform traditional statistical methods by capturing complex, non-linear interactions among variables [37]. Deep learning approaches, including convolutional neural networks (CNNs) and recurrent neural networks (RNNs), further enhance prediction accuracy by analyzing sequential learning data and identifying temporal patterns in student behavior [38].

In parallel, learning analytics combined with big data methods has become central to performance forecasting. Clickstream data collected from learning management systems—tracking time on task, navigation behavior, and interaction frequency—provides valuable insights when analyzed through AI algorithms, allowing for the early identification of at-risk students using explainable AI [39]. Integrating multimodal data, such as academic performance, demographic information, social networks, and behavioral signals, leads to more accurate and comprehensive predictive models [40].

AI also facilitates real-time prediction and intervention through early warning systems that detect students at risk of failure or dropout, often achieving over 80% prediction accuracy, allowing for continuous monitoring throughout the term [41]. Adaptive learning platforms, driven by intelligent tutoring systems, utilize predictive analytics to tailor learning content to individual student profiles, dynamically adjusting the difficulty and pacing to optimize learning outcomes [42].

Natural language processing (NLP) has further expanded the role of AI in education. By analyzing student-generated text from essays, discussion forums, and feedback, NLP tools can evaluate performance and engagement. Large language models, such as GPT and BERT, have demonstrated effectiveness in automated essay scoring and the semantic understanding of student writing, offering scalable solutions for qualitative assessment [43].

*2.6. AI-Based Prediction of Student Performance Using Personality Traits*

The integration of personality data into AI-based predictive models marks a significant development in educational data mining. Personality traits offer stable, theory-based variables that enhance the predictive power of traditional academic and behavioral indicators. Feature engineering efforts increasingly include the Big Five personality dimensions, which have been shown to improve model accuracy by 5–15% compared to models relying solely on academic and demographic data [44]. Furthermore, AI can now infer personality traits from digital behaviors such as social media activity, learning management system interactions, and textual data, using natural language processing to extract personality indicators with reasonable accuracy [45].

Methodologically, the use of ensemble methods—combining various machine learning algorithms—has led to improved predictive performance when personality data is included. These methods balance the strengths of different algorithms, enhancing the robustness of predictions [46]. Deep learning models have also incorporated personality traits into their architectures, allowing for the capture of complex, non-linear relationships between these traits and other student characteristics, thereby improving prediction outcomes [47].

The predictive performance of models that integrate personality data is strong, with reported accuracy rates ranging from 75% to 90% for various academic outcomes. Including personality traits typically enhances evaluation metrics such as precision, recall, and F1-scores [48]. These models also show good temporal stability, maintaining accuracy across different academic terms and cohorts due to the consistency of personality traits over time [49].

However, the use of personality data introduces significant ethical challenges. Concerns regarding privacy, informed consent, and algorithmic bias must be addressed to ensure the responsible use of this data. Research highlights the importance of transparent algorithms, opt-in consent mechanisms, and ongoing bias audits to safeguard ethical standards in the educational applications of personality-driven AI systems [50].

**Research Gap: What Specific Knowledge Gap Does This Study Address?**

While leadership traits and personality have been linked to academic success, there is limited research exploring the direct correlation between leadership personality traits and students' average grades.

- Most studies focus on general personality dimensions (Big Five traits) rather than leadership-specific traits such as transformational leadership behaviors, emotional intelligence competencies, and social influence capabilities.
- Limited integration of machine learning approaches exists to predict student performance based on comprehensive personality assessments. Current AI-based prediction models predominantly utilize demographic, behavioral, and basic academic variables, while underutilizing the predictive power of leadership-specific personality traits.
- There is a need for more multi-dimensional analyses, combining self-assessment results with academic data across diverse student populations. Most existing research relies on single-source data collection methods rather than comprehensive assessment frameworks that integrate multiple personality measurement approaches.
- Insufficient focus on leadership trait specificity: While general personality research in academic contexts is extensive, there remains a significant gap in understanding how specific leadership characteristics (such as inspirational motivation, intellectual stimulation, individualized consideration, and idealized influence) directly correlate with academic performance outcomes.
- Lack of comprehensive machine learning modelling: Few research models systematically integrate advanced machine learning techniques with leadership personality assessments to create robust predictive frameworks for academic success.
- Limited multi-dimensional analytical approaches: Current research typically examines isolated relationships between single personality dimensions and academic outcomes, rather than exploring complex interactions and patterns through comprehensive analytical frameworks.

This study aims to address these gaps by using a comprehensive personality assessment and machine learning modelling approach, offering more profound insights into how leadership traits influence academic achievement through:
1. Direct examination of leadership-specific personality traits and their correlation with student GPA.
2. Integration of advanced machine learning algorithms to create predictive models based on leadership personality assessments.
3. Multi-dimensional analysis combining self-assessment data with academic performance across diverse student populations.
4. Development of a comprehensive framework that bridges leadership theory, personality psychology, and educational data science.

**Related Work Summary**

This exploration of the state-of-the-art literature reveals a rich landscape of research encompassing leadership trait theory, determinants of academic performance, relationships between personality and achievement, self-assessment methodologies, and AI-based educational prediction systems. The convergence of these fields presents significant opportunities for advancing our understanding of student success and developing more effective predictive models.

The identified research gaps highlight the need for more integrated approaches that combine leadership trait theory with advanced AI methodologies, consider cultural and contextual factors, and focus on actionable insights for educational practice. Future research should prioritize longitudinal studies that examine the dynamic relationships between personality, leadership development, and academic success, while also developing ethical frameworks for AI-based prediction systems in educational contexts.

The integration of personality traits, particularly leadership-relevant characteristics, into AI-based prediction systems represents a promising avenue for enhancing student success prediction and intervention. However, realizing this potential requires addressing the methodological, ethical, and practical challenges identified in this review while maintaining focus on the ultimate goal of improving educational outcomes for all students.

## 3. Materials and Methods

For the research design, to achieve the objective of this study, we employed a quantitative approach, which involved the use of techniques such as Exploratory Data Analysis, Correlation Analysis and Machine Learning (ML) techniques to determine academic success based on personality variables quantified through several personality tests.

129 samples of MSc student assessments of the leadership module collected in the Environmental Engineering Department. The sample is geographically diverse, comprising students from 17 countries. The male/female ratio is 1.3/1. Twenty-three (23) personal traits were investigated through self-analytical tools, such as Personality Insight, Workplace Culture, Motivation at Work, Management Skills, and Emotion Control. Personality Inside test checks for extraversion, agreeableness, emotional stability, conscientiousness, and openness to experience. Workplace Culture tests for immediate, entrepreneurial/creative, family, achievement, mission and bureaucratic culture. Motivation at Work focuses on development, purpose, control, recognition, status, failure aversion, reward, Achievement, stability, and interaction. Academic performance metrics (averaging students' performance) were collected from academic reports. The feature description used in the study is provided in Table 1.

**Table 1.** Variable/feature names and their characteristics.

| Variable Names | Characteristics |
|---|---|
| **Motivation at Work** | |
| Development | The need for personal and professional growth, learning, and improvement. |
| Purpose | The desire to do meaningful work that aligns with personal values or contributes to a greater good. |
| Control | The need for autonomy and the ability to influence decisions that affect one's work. |
| Recognition | The need to be acknowledged or praised for contributions and achievements. |
| Status | The motivation to attain prestige, rank, or reputation in an organization or field. |
| Failure Aversion | Motivation driven by a fear of failure often leads to over-preparation or risk avoidance. |
| Reward | Motivation is based on tangible benefits, such as salary, bonuses, perks, or other external incentives. |
| Achievement | A drive to excel, master tasks, and accomplish challenging goals. |
| Stability | The desire for predictability, job security, and consistency. |
| Interaction | The motivation that comes from social relationships and teamwork. |
| **Personality Inside** | |

| Extraversion | The extent to which someone is outgoing, energetic, and sociable. |
| Agreeableness | The degree of a person's compassion, cooperativeness, and consideration toward others. |
| Emotional Stability | The ability to remain calm, resilient, and free from persistent negative emotions. |
| Conscientiousness | Reflects how organized, responsible, and goal-oriented a person is. |
| Openness to Experience | Describes how creative, curious, and open-minded someone is. |
| **Workplace Culture** | |
| Immediate Culture | A fast-paced, results-driven environment focused on quick decisions and rapid responses. |
| Entrepreneurial / Creative Culture | A culture that encourages innovation, experimentation, and risk-taking. |
| Family Culture | A collaborative and supportive environment that emphasizes loyalty, trust, and a sense of belonging. |
| Achievement Culture | Focused on high performance, competition, and goal achievement. |
| Bureaucratic Culture | A structured, rules-based environment with clear procedures and hierarchy. |
| Mission Culture | Driven by a strong sense of purpose, values, and organizational goals. |
| **Management Skills** | An individual's ability to lead, organise, make decisions, communicate effectively, and manage people and resources effectively. |
| **Emotional Control** | An individual's ability to recognise, manage, and respond to emotional experiences, especially under stress or pressure, is often linked to Emotional Intelligence (EQ). |

*3.1. Dataset preparation procedures*

After data retrieval, the results from the self-assessment test were scaled between 1 (low) and 5 (high) and visualised in histograms. Management skills results were ranked between 0 and 130, and the emotional control tests were ranked between 0 and 100. Those results, together with the Leadership in the Digital Age mark, were matched to the average grades of each student. The final dataset comprised 129 participants and included 25 variables.

Written consent regarding the use of data was obtained from the students. The dataset preparation procedure adhered to ethical considerations and data privacy protection policies. Personal information, such as student names and student IDs, was anonymized.

*3.2. Data analysis and modelling methods*

The tests were done on Python. For the results' visualization were used the following libraries: seaborn and matplotlib.

Feature selection was based on the values of their correlation coefficients. Sixteen features with the coefficients 0.05 or above were selected for machine learning modelling.

The average grades were separated into three categories: "Fail" with a score below 50 out of 100, "Pass" with a score between 50 and 70, and "Excellent" with a score above 70.

Seven ML algorithms, such as Support Vector Machine (SVM), Logistic Regression (LR), K-Nearest Neighbours (KNN), Decision Tree (DT), Gradient Boosting (GB), Random Forest (RF), XGBoost (XGB) and LightGBM (LGBM), were tested in the study.

Hyperparameter search grid (GridSearchCV) was used to optimize the performance of each based model. The procedure was accompanied with 5-fold cross-validation. The final best grids used for models' training are summarized in Table 2.

**Table 2.** Optimal hyperparameters are used for modelling.

| ML Algorithm | Hyperparameters |
|---|---|
| SVM | {'C': 10, 'kernel': 'rbf'} |
| LG | {'C': 10, 'penalty': 'l2', 'solver': 'lbfgs'} |
| KNN | {'n_neighbors': 5, 'weights': 'distance'} |
| DT | {'max_depth': None, 'min_samples_split': 5} |
| GB | {'max_depth': 7, 'n_estimators': 200} |
| RF | {'max_depth': 10, 'min_samples_split': 2, 'n_estimators': 200} |
| XGB | {'max_depth': None, 'min_samples_split': 2, 'n_estimators': 200} |
| LGBM | {'max_depth': None, 'min_samples_split': 2, 'n_estimators': 200} |

## 4. Results

*4.1. Descriptive statistics and correlation analysis*

The dataset was checked for missing, duplicate values, and outliers. A clean dataset was used for the correlation analysis of personal leadership traits and academic performance. Figure 1 reflects scores of 23 normalized personal traits after data standardization using a standard scaler.

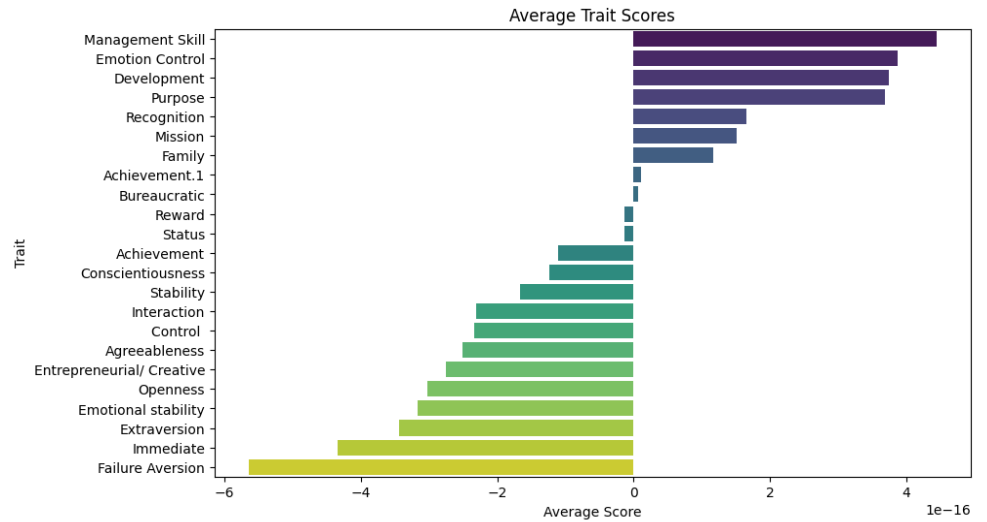

**Figure 1.** Personal traits scaling after standardization procedure. This horizontal bar chart shows average trait scores across various personality and behavioral characteristics. Management Skill, Emotion Control, and Development rank highest with positive scores around 4-5, while middle-tier traits like Purpose and Recognition show moderate positive scores. Lower-scoring traits include Agreeableness and Emotional Stability, with Failure Aversion showing the most negative score at approximately -4. The chart uses a color gradient from purple (highest) to yellow (lowest) to distinguish performance levels.

Figure 2 illustrates the correlation between individual leadership traits and leadership performance ratings.

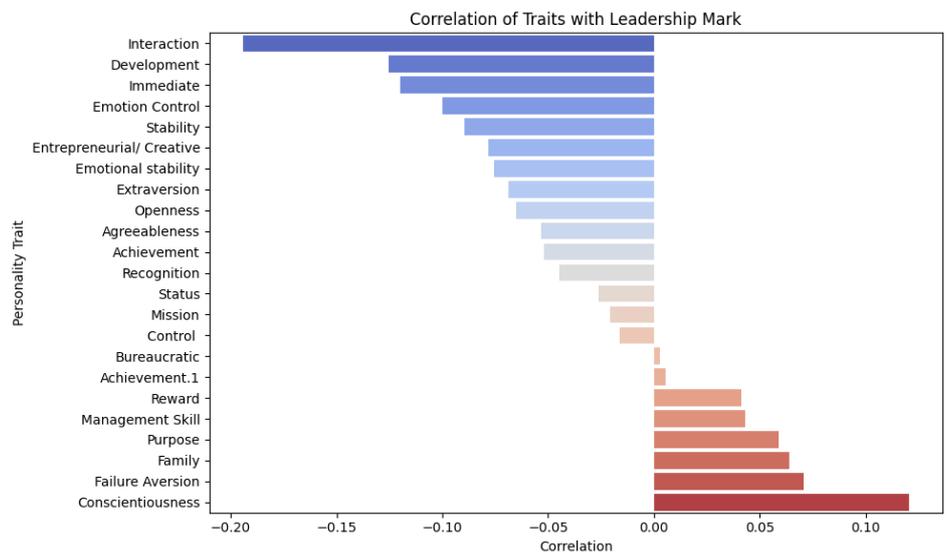

**Figure 2.** Correlation dependencies between personal traits and leadership mark. This horizontal bar chart displays the correlation of various personality traits with Leadership Mark. Traits are ordered from highest positive correlation (Interaction, Development, Immediate) at the top in blue, through neutral correlations in the middle, to negative correlations at the bottom in red. Conscientiousness and Failure Aversion show the strongest negative correlations with leadership, while traits like Management Skill, Purpose, and Family also correlate negatively. The correlation values range from approximately -0.20 to +0.15.

The highest correlation coefficients (below -0.1 and above 0.1) were identified between leadership marks and variables such as "interaction", "development", "immediacy", "emotional control", and "conscientiousness". It is important to note that among the mentioned features, only "conscientiousness" shows a positive correlation with the mark. Similarly, the correlation between the average grade and personal traits was checked. The results of it are reflected in Figure 3.

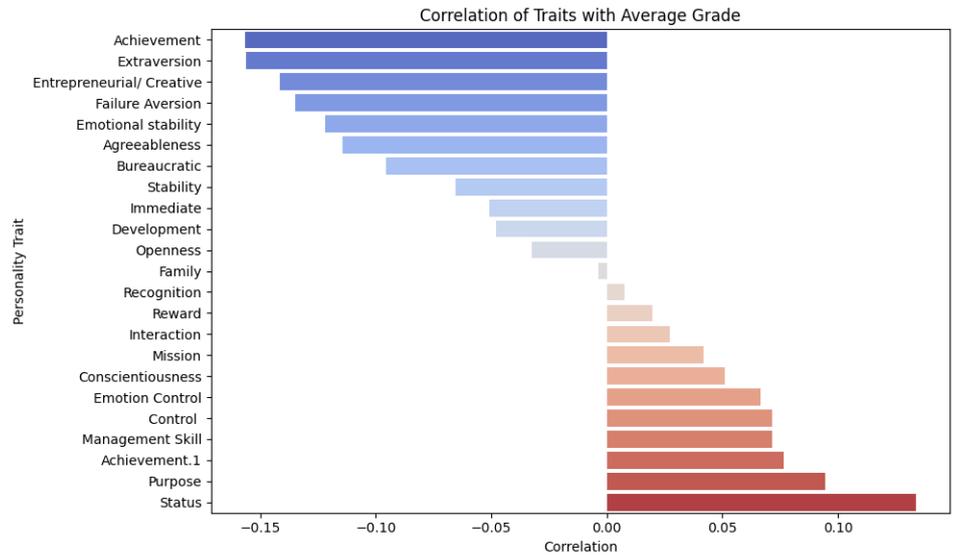

**Figure 3.** Correlation dependencies between personal traits and average grade. This horizontal bar chart shows the correlation of personality traits with Average Grade. Achievement, Extraversion, and Entrepreneurial/Creative traits show the strongest positive correlations with grades (around 0.10-0.15) in blue. Most traits in the middle show weak or near-zero correlations. At the bottom, traits like Purpose, Achievement-1, Management Skill, and Status display negative correlations with academic performance in red, with Status showing the strongest negative correlation at approximately -0.10.

The strongest correlation coefficients were identified between (below -0.1 and above 0.1) "achievement", "extraversion", "entrepreneurial/creative", "failure aversion", "emotional stability", "agreeableness", and "status". The first six features in the row have a negative correlation with performance, and the only "status" feature demonstrates a positive effect on performance. The correlation coefficients for each feature are given in Table 3.

**Table 3.** Feature correlation coefficients.

| Variable Name | Pearson Correlation Coefficient |
| --- | --- |
| Averages | 1.000000 |
| Leadership Mark | 0.265457 |
| Status | 0.133898 |
| Purpose | 0.094405 |
| Achievement.1 | 0.076527 |
| Management Skills | 0.071644 |
| Control | 0.071583 |
| Emotion Control | 0.066421 |
| Conscientiousness | 0.051116 |
| Mission | 0.041769 |
| Integration | 0.027280 |
| Reward | 0.019607 |
| Recognition | 0.007730 |
| Family | -0.003501 |
| Openness | -0.032390 |
| Development | -0.048041 |
| Immediate | -0.050725 |
| Stability | -0.065290 |
| Bureaucratic | -0.095548 |
| Agreeableness | -0.114376 |
| Emotional Stability | -0.122030 |
| Failure Aversion | -0.134780 |
| Entrepreneurial/Creative | -0.141348 |
| Extraversion | -0.156131 |
| Achievement | -0.156720 |

*4.2. Feature selection and dataset balancing*

The feature selection procedure was based on feature filtering, using a correlation coefficient threshold below -0.05 and above 0.05. 18 features were selected for the modelling.

Before data modelling, a new column called "Performance" was created. The "Performance" column included three groups (classes) of the results based on averaging values: "Fail", "Pass" and "Excellent". The "Averages" variable was dropped. After that, the classes of the dataset were balanced using the Synthetic Minority Over-sampling Technique (SMOTE). The original data has a shape of 129x18, while the resampled data has a shape of 276x18. Class distribution before SMOTE: class 0 – 12 samples, class 1 – 25 samples, class 2 – 92 samples. Class distribution after SMOTE: class 0 – 92 samples, class 1 – 92 samples, class 2 – 92 samples.

*4.3. Data modelling*

The data was separated into training, validation, and testing parts in a proportion of 80:10:10. Table 4 summarises the machine learning (ML) performance metrics for the testing dataset of SVM, LR, KNN, DT, GB, RF, XGB, and LGBM.

**Table 4.** ML models classification performance for three categories of grades.

| Model | Accuracy | Precision | Recall | F1-Score |
|---|---|---|---|---|
| Random Forest | 0.875000 | 0.893856 | 0.875000 | 0.875662 |
| XGBoost | 0.875000 | 0.875642 | 0.875000 | 0.872126 |
| Gradient Boosting | 0.857143 | 0.882311 | 0.857143 | 0.856874 |
| LightGBM | 0.857143 | 0.866562 | 0.857143 | 0.855138 |
| Support Vector Machine | 0.839286 | 0.868452 | 0.839286 | 0.832172 |
| Logistic Regression | 0.821429 | 0.821429 | 0.821429 | 0.821429 |
| Decision Tree | 0.767857 | 0.777381 | 0.767857 | 0.770364 |
| K- Nearest Neighbours | 0.696429 | 0.672958 | 0.696429 | 0.656341 |

Overall, the best model for the current dataset is Random Forest, which demonstrates 87.50% accuracy, 89.38% precision, 87.50% recall, and 87.56% F1-score. Very close to RF performance, XGBoost achieves an accuracy of 87.50%, precision of 87.56%, recall of 87.50%, and F1-score of 87.21 (see Figure 4). The less effective models in classification tasks are DT and KNN.

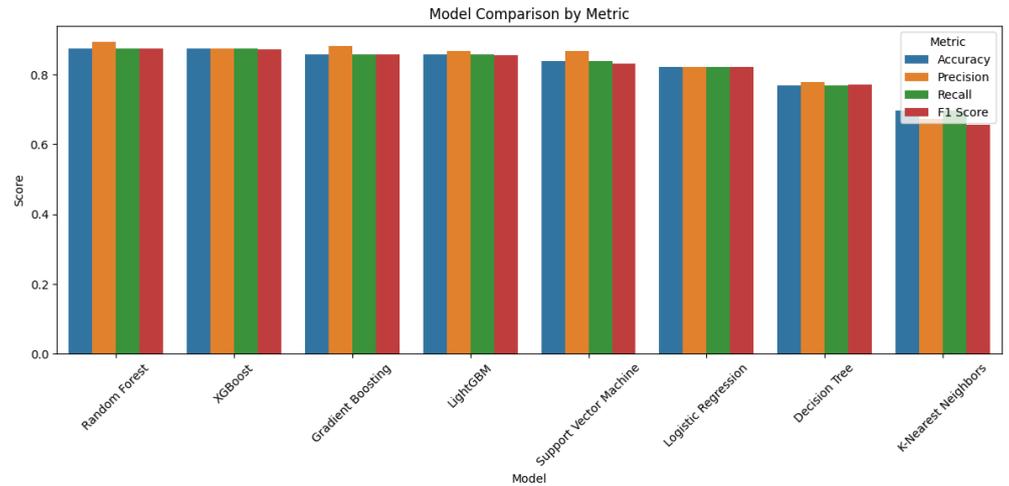

**Figure 4.** Performance metrics in comparison. This grouped bar chart compares the performance of nine machine learning models across four evaluation metrics: Accuracy, Precision, Recall, and F1 Score. Most models achieve scores between 0.7-0.9, with Random Forest, XGBoost, and Gradient Boosting showing the highest performance across all metrics. Decision Tree and K-Nearest Neighbor display the lowest scores, around 0.6-0.7. The metrics are represented by different colored bars (blue, orange, green, and red) for each model, showing relatively consistent performance patterns across the different evaluation measures.

Additional test for dataset modelling was conducted without the "Leadership Mark" variable. Table 5 demonstrates the results of modelling based on 17 personal traits features for the testing set.

**Table 5.** Model's performance for test for 17 leadership traits and 3 categories of grades.

| Model | Accuracy | Precision | Recall | F1-Score |
| --- | --- | --- | --- | --- |
| Random Forest | 0.857143 | 0.879104 | 0.857143 | 0.857560 |
| XGBoost | 0.821429 | 0.828508 | 0.821429 | 0.823872 |
| LightGBM | 0.803571 | 0.818027 | 0.803571 | 0.805883 |
| Support Vector Machine | 0.803571 | 0.828125 | 0.803571 | 0.782710 |
| Gradient Boosting | 0.767857 | 0.773101 | 0.767857 | 0.770135 |
| Decision Tree | 0.750000 | 0.751284 | 0.750000 | 0.748035 |
| Logistic Regression | 0.732143 | 0.719345 | 0.732143 | 0.719239 |
| K- Nearest Neighbours | 0.714286 | 0.695727 | 0.714286 | 0.680463 |

The best model, similar to the first test, is RF, with an accuracy of 85.71%, a precision of 87.91%, a recall of 85.71%, and an F1-score of 85.75%. The XGB model remains in second place with an accuracy of 82.14%, precision of 82.85%, recall of 82.14%, and F1-score of 82.38%. The worst-performing models are LR and KNN. Figure 5 illustrates performance metrics in comparison.

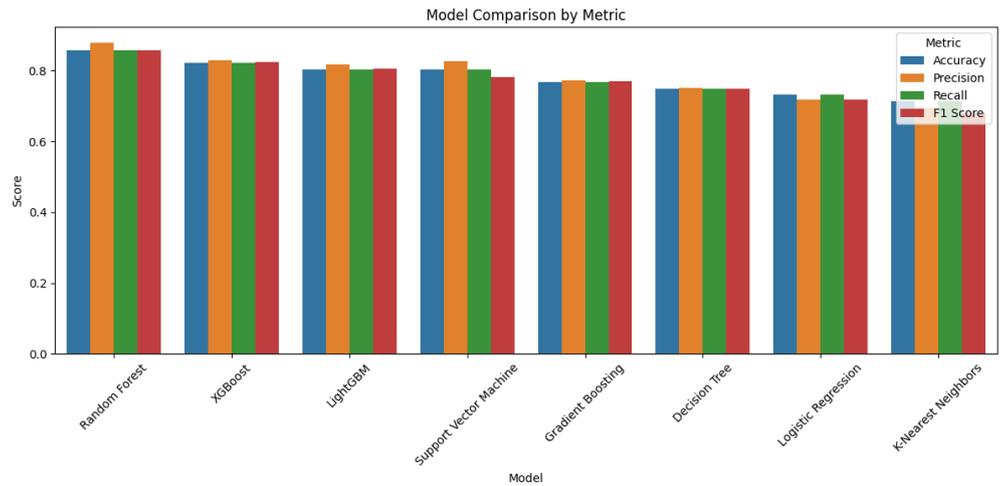

**Figure 5.** Performance metrics in comparison for 17 features and 3 classes. This grouped bar chart compares the performance of eight machine learning models across four evaluation metrics: Accuracy, Precision, Recall, and F1 Score. Random Forest and XGBoost achieve the highest performance with scores around 0.85-0.9 across all metrics. Most models perform in the 0.7-0.8 range, while Decision Tree, Logistic Regression, and K-Nearest Neighbor show the lowest scores around 0.65-0.75. The four metrics are represented by different colored bars (blue, orange, green, and red), with most models showing consistent performance across all evaluation measures.

All the performance results were validated using a 5 cross-validation procedure.

*4.4. Feature Importance*

The feature importance for the RF, XGB, LGBM, GB, DT, and LR models is shown in Figure 6.

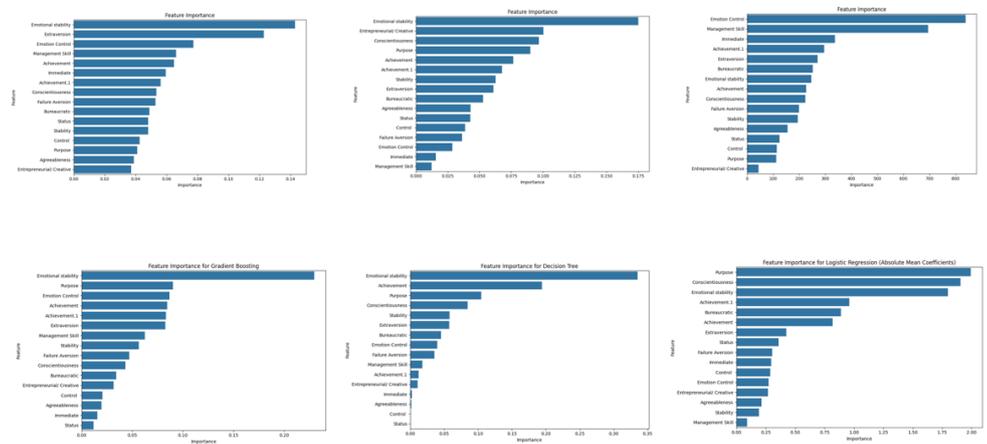

**Figure 6.** Feature importance for machine learning modelling. This figure displays six horizontal bar charts showing feature importance rankings across different machine learning models or analyses. Each subplot contains multiple features ranked by their importance scores, with longer blue bars indicating higher importance. The charts appear to compare how different features contribute to model performance, with each panel showing a distinct ranking pattern. The features are listed on the y-axis while importance values are shown on the x-axis, allowing for easy comparison of which variables are most influential in each respective model or analysis.

For SVM and KNN models, feature importance was not directly available.

Table 6 highlights the five most essential features in the models.

Table 6. Feature importance for modelling.

| RF | XGBoost | LGBM | GB | DT | LR |
|---|---|---|---|---|---|
| Emotional Stability | Emotional Stability | Emotional Control | Emotional Stability | Emotional Stability | Purpose |
| Extraversion | Entrepreneurial/ Creative | Management Skill | Purpose | Achievement | Conscientiousness |
| Emotion Control | Conscientiousness | Immediate | Emotion Control | Purpose | Emotional Stability |
| Management Skill | Purpose | Achievement.1 (Achievement Culture) | Achievement | Conscientiousness | Achievement.1 (Achievement Culture) |
| Achievement | Achievement | Extraversion | Achievement.1 (Achievement Culture) | Stability | Bureaucratic |

Feature importance indicates that emotional stability is the most valuable feature, as it appears in all 6 models in the table. The second most frequent features are achievement, achievement culture, and purpose (3 out of 6), followed by emotional control and conscientiousness (2 out of 6).

## 5. Discussion and Conclusions

The study demonstrates the high predictive power of AI technologies for average students' grades based on their leadership traits. The study presents an additional opportunity to identify students' strengths and weaknesses at an early stage and choose the correct strategies for personalized learning.

Compared to the existing literature, which has primarily focused on general personality dimensions (e.g., Big Five traits) and their impact on students' performance [44, 45], this study examines the influence of leadership traits. Most of the existing research relies on single-source data. In contrast, the current study combined five leadership personality tests. The state-of-the-art literature involving AI technology reflects behavioral patterns [37, 38, 40] or relies on text analysis for grade prediction [43]. A minimal number of studies describe the predictions of student grades based on personality types, and none of them were focused on the performance predictive modelling using leadership traits.

Eighteen for the first model and seventeen personality traits for the second model were used in the current research. The highest training and testing performances were achieved with RF, with an accuracy of 87.50% for the model that included leadership mark for the model building, and an accuracy of 85.71% for the model without it. The accuracy rates for academic outcome prediction using heterogeneous data vary between 75% and 90% in the state-of-the-art literature. From this perspective, our predictions are highly accurate and have the potential to be further improved.

The research helped provide answers to the research questions.

1. How do leadership personality traits correlate with students' academic performance?

The students' academic performance shows a direct linear correlation between leadership personality traits and average grades.

2. Which leadership personality traits appear most influential in predicting higher student grades?

The strongest correlation coefficients were identified between the average grade and traits such as achievement, extraversion, entrepreneurial/creative culture, failure aversion, emotional stability, agreeableness, and motivation to attain a status. At the same time, the most influential features in machine learning modelling were leadership traits, including a drive to accomplish challenging goals (achievement), a focus on high performance, competition, and goal achievement (achievement culture), a desire to do meaningful work (purpose), emotional control, and conscientiousness. Most of all, it is essential to note that some of the features have a positive impact on the average performance, and some have an adverse effect. We expect that feature importance can vary between university departments and reflect the differences between students in STEM and humanities subjects. Further research can give a deeper insight into personality traits and their effect on student achievements.

3. Can a machine learning model accurately predict students' grade categories based on leadership personality traits?

Three categories of performance, "Pass", "Excellent", and "Fail", were identified using a Random Forest model with an accuracy of 85.71%, precision of 87.91%, recall of 85.71% and F1-score of 85.75%.

4. How can universities use insights from this study to develop personalized academic strategies tailored to students' strengths and personality profiles?

Comprehensive leadership trait development strategies can be incorporated into university programs. For example, building an intrapreneurial culture and developing a teaching strategy on how to accomplish challenging goals and develop emotional stability during the project development cycle. At the same time, students who have insufficient or undeveloped leadership traits can create a personal development plan to address weaknesses and strengthen traits that contribute to higher marks. Students can be offered stress management programs. Additionally, students with a high probability of failing should receive extra academic and psychological support tailored to their personality types.

The suggested strategies can be based on student trait profiles. For example, an introverted student with a high level of conscientiousness will be assigned to an independent research or project-based task. A student with a high level of openness, who prefers a low-structured environment, will be recommended interdisciplinary electives and creative assessments. A highly extraverted person with low self-discipline will be allocated to collaborative study groups with structured timeframes. Finally, a student with high neuroticism and low resilience will be paired with a mentor and included in the stress management workshops.

The course of study will be offered, structured and dynamically adjusted using properties of ML models and students' personality features. Course recommendations will be tuned according to the predicted difficulty. A personality profile might help with selecting the assignment types (e.g., projects vs. timed tests) and with instruction formats (e.g., visual vs. audio vs. textual content).

Despite promising findings of the current research, data collection was limited by the EE department (e.g., business management, humanities, and other departments were not included in the study at this stage). The sample size could also be extended.

Future research directions could utilize additional features obtained from career development tests (e.g., personal resilience, learning style, or sound decision-making) and, together with leadership traits, explore their effects on students' performance.

Modelling of student performance, including that of other university departments and programs, can provide a deeper understanding of human nature and its impact on student achievements.

While developing new programs or platforms for personalized learning [51], the researchers and software developers must consider the complexity of the program and potential user issues with AI tools [52].

Future research will focus on developing a teaching/learning framework that helps students personalize their learning and develop valuable skills for achieving high academic performance.

**Supplementary Materials:** Not available.

**Author Contributions:** Conceptualization, N.H. and R.F.; methodology, N.H.; software, N.H.; validation, N.H., D.H. and R.B.S; formal analysis, N.H.; investigation, N.H.; resources, N.H.; data curation, N.H.; writing—original draft preparation, N.H., R.B.S., and D.H; visualization, N.H.; project administration, N.H. All authors have read and agreed to the published version of the manuscript.

**Funding:** This research received no external funding.

**Data Availability Statement:** The data is held on private university domain.

**Conflicts of Interest:** The authors declare no conflicts of interest.

# References

1. Luckin, R. and Cukurova, M., 2019. Designing educational technologies in the age of AI: A learning sciences-driven approach. *British Journal of Educational Technology*, 50(6), pp.2824-2838.
2. Ouyang, F. and Zhang, L., 2024. AI-driven learning analytics applications and tools in computer-supported collaborative learning: A systematic review. *Educational Research Review*, 44, p.100616.
3. Vorobyeva, K.I., Belous, S., Savchenko, N.V., Smirnova, L.M., Nikitina, S.A. and Zhdanov, S.P., 2025. Personalized Learning through AI: Pedagogical Approaches and Critical Insights. *Contemporary Educational Technology*, 17(2).
4. Halkiopoulos, C. and Gkintoni, E., 2024. Leveraging AI in e-learning: Personalized learning and adaptive assessment through cognitive neuropsychology—A systematic analysis. *Electronics*, 13(18), p.3762.
5. Carlyle, T., 1993. *On heroes, hero-worship, and the heroic in history* (Vol. 1). Univ of California Press.
6. Organ, D.W., 1996. Leadership: The great man theory revisited. *Business horizons*, 39(3), pp.1-4.
7. Germain, M.-L. (2008). Traits and skills theories as the nexus between leadership and expertise: Reality or fallacy? *Paper presented at the Academy of Human Resource Development International Research Conference in the Americas, Panama City, FL, February 20–24*. Available at: https://files.eric.ed.gov/fulltext/ED501636.pdf.
8. Kirkpatick, S.A. and Locke, E.A., 1991. Leadership: do traits matter? *Academy of management perspectives*, 5(2), pp.48-60.
9. Northouse, P. G. (2021). Leadership: Theory and practice (8th ed.). *Sage Publications*.
10. Judge, T. A., Colbert, A. E., & Ilies, R. (2004). Intelligence and leadership: A quantitative review and test of theoretical propositions. *Journal of Applied Psychology*, 89(3), 542-552.
11. Goleman, D. (1995). Emotional intelligence. *Bantam Books*.
12. Judge, T. A., Bono, J. E., Ilies, R., & Gerhardt, M. W. (2002). Personality and leadership: A qualitative and quantitative review. *Journal of Applied Psychology*, 87(4), 765-780.
13. McClelland, D. C. (1985). Human motivation. *Scott, Foresman*.
14. Peterson, C., & Seligman, M. E. (2004). Character strengths and virtues: A handbook and classification. *Oxford University Press*.
15. Avolio, B. J., & Gardner, W. L. (2005). Authentic leadership development: Getting to the root of positive forms of leadership. *The Leadership Quarterly*, 16(3), 315-338.
16. Van Dierendonck, D. (2011). Servant leadership: A review and synthesis. *Journal of Management*, 37(4), 1228-1261.
17. Richardson, M., Abraham, C., & Bond, R. (2012). Psychological correlates of university students' academic performance: A systematic review and meta-analysis. *Psychological Bulletin*, 138(2), 353-387.
18. Poropat, A. E. (2014). Other-rated personality and academic performance: Evidence and implications. *Learning and Individual Differences*, 34, 24-32.
19. Zimmerman, B. J. (2002). Becoming a self-regulated learner: An overview. Theory Into Practice, 41(2), 64-70.
20. Dweck, C. S. (2006). Mindset: The new psychology of success. *Random House*.

21. Sirin, S. R. (2005). Socioeconomic status and academic achievement: A meta-analytic review of research. *Review of Educational Research*, 75(3), 417-453.
22. Tinto, V. (1993). Leaving college: Rethinking the causes and cures of student attrition (2nd ed.). *University of Chicago Press*.
23. Pascarella, E. T., & Terenzini, P. T. (2005). How college affects students: A third decade of research. *Jossey-Bass*.
24. Trapmann, S., Hell, B., Hirn, J. O. W., & Schuler, H. (2007). Meta-analysis of the relationship between the Big Five and academic success at university. *Zeitschrift für Psychologie*, 215(2), 132-151.
25. Poropat, A. E. (2009). A meta-analysis of the five-factor model of personality and academic performance. *Psychological Bulletin*, 135(2), 322-338.
26. Chamorro-Premuzic, T., & Furnham, A. (2003). Personality predicts academic performance: Evidence from two longitudinal university samples. *Journal of Research in Personality*, 37(4), 319-338.
27. Bandura, A. (1997). Self-efficacy: The exercise of control. *Freeman*.
28. Duckworth, A. L., Peterson, C., Matthews, M. D., & Kelly, D. R. (2007). Grit: Perseverance and passion for long-term goals. *Journal of Personality and Social Psychology*, 92(6), 1087-1101.
29. John, O. P., & Srivastava, S. (1999). The Big Five trait taxonomy: History, measurement, and theoretical perspectives. *Handbook of Personality: Theory and Research*, 2, 102-138.
30. Roberts, B. W., & DelVecchio, W. F. (2000). The rank-order consistency of personality traits from childhood to old age: A quantitative review of longitudinal studies. *Psychological Bulletin*, 126(1), 3-25.
31. Connolly, J. J., Kavanagh, E. J., & Viswesvaran, C. (2007). The convergent validity between self and observer ratings of personality: A meta-analytic review. *International Journal of Selection and Assessment*, 15(1), 110-117.
32. Paulhus, D. L. (2002). Socially desirable responding: The evolution of a construct. *The Role of Constructs in Psychological and Educational Measurement*, 49-69.
33. Taylor, S. E., & Brown, J. D. (1988). Illusion and well-being: A social psychological perspective on mental health. *Psychological Bulletin*, 103(2), 193-210.
34. Dunning, D., Heath, C., & Suls, J. M. (2004). Flawed self-assessment: Implications for health, education, and the workplace. *Psychological Science in the Public Interest*, 5(3), 69-106.
35. Andrade, H., & Du, Y. (2007). Student responses to criteria-referenced self-assessment. *Assessment & Evaluation in Higher Education*, 32(2), 159-181.
36. Panadero, E., Jonsson, A., & Botella, J. (2017). Effects of self-assessment on self-regulated learning and self-efficacy: Four meta-analyses. *Educational Research Review*, 22, 74-98.
37. Romero, C., & Ventura, S. (2020). Educational data mining and learning analytics: An updated survey. *Wiley Interdisciplinary Reviews: Data Mining and Knowledge Discovery*, 10(3), e1355.
38. Ujkani, B., Minkovska, D. and Hinov, N., 2024. Course success prediction and early identification of at-risk students using explainable artificial intelligence. *Electronics*, 13(21), p.4157.
39. Chen, X., Zou, D., Cheng, G., & Xie, H. (2020). Detecting latent topics and trends in educational technologies over four decades using structural topic modelling. *Computers & Education*, 151, 103846.
40. Dutt, A., Ismail, M. A., & Herawan, T. (2017). A systematic review on educational data mining. *IEEE Access*, 5, 15991-16005.
41. Helal, S., Li, J., Liu, L., Ebrahimie, E., Dawson, S., Murray, D. J., & Long, Q. (2018). Predicting academic performance by considering student heterogeneity. *Knowledge-Based Systems*, 161, 134-146.
42. Chrysafiadi, K., & Virvou, M. (2013). Student modelling approaches: A literature review for the last decade. *Expert Systems with Applications*, 40(11), 4715-4729.
43. Shermis, M. D., & Burstein, J. (Eds.). (2013). Handbook of automated essay evaluation: Current applications and new directions. *Routledge*.
44. Okubo, F., Yamashita, T., Shimada, A., & Ogata, H. (2017). A neural network approach for students' performance prediction. *Proceedings of the Seventh International Learning Analytics & Knowledge Conference*, 598-599.
45. Park, G., Schwartz, H. A., Eichstaedt, J. C., Kern, M. L., Kosinski, M., Stillwell, D. J., ... & Seligman, M. E. (2015). Automatic personality assessment through social media language. *Journal of Personality and Social Psychology*, 108(6), 934-952.
46. Ahmad, F., Ismail, N. H., & Aziz, A. A. (2015). The prediction of students' academic performance using classification data mining techniques. *Applied Mathematical Sciences*, 9(129), 6415-6426.
47. Hussain, M., Zhu, W., Zhang, W., & Abidi, S. M. R. (2018). Student engagement predictions in an e-learning system and their impact on student course assessment scores. *Computational Intelligence and Neuroscience*, 2018.

48. Karagiannis, I., & Satratzemi, M. (2018). An adaptive mechanism for Moodle based on automatic detection of learning styles. *Education and Information Technologies*, 23(3), 1331-1357.
49. Kosinski, M., Matz, S. C., Gosling, S. D., Popov, V., & Stillwell, D. (2015). Facebook as a research tool for the social sciences. *American Psychologist*, 70(6), 543-556.
50. Baker, R. S., & Hawn, A. (2022). Algorithmic bias in education. *International Journal of Artificial Intelligence in Education*, 32(4), 1052-1092.
51. Schicchi, D. and Taibi, D., 2024, November. Redefining education: A personalized ai platform for enhanced learning experiences. *In Proceedings of the Second International Workshop on Artificial Intelligence Systems in Education co-located with 23rd International Conference of the Italian Association for Artificial Intelligence (AIxIA 2024), volume—of CEUR Workshop Proceedings.*
52. Gunawardena, M., Bishop, P. and Aviruppola, K., 2024. Personalized learning: The simple, the complicated, the complex and the chaotic. *Teaching and Teacher Education*, 139, p.104429.